\documentclass[letterpaper]{article} 
\usepackage{aaai25}  
\usepackage{times}  
\usepackage{helvet}  
\usepackage{amsmath}
\usepackage{multirow}
\usepackage{tabularx}
\usepackage{booktabs}
\usepackage{multicol}
\usepackage{enumitem}
\usepackage{subcaption}
\usepackage{xcolor}
\definecolor{shadecolor}{rgb}{0.92,0.92,0.92}
\usepackage{courier}  
\usepackage[hyphens]{url}  
\usepackage{graphicx} 
\usepackage{natbib}  
\usepackage{caption} 
\usepackage{algorithm}
\usepackage{algorithmic}

\usepackage{newfloat}
\usepackage{listings}
\DeclareCaptionStyle{ruled}{labelfont=normalfont,labelsep=colon,strut=off} 
\floatstyle{ruled}
\newfloat{listing}{tb}{lst}{}
\floatname{listing}{Listing}
\title{Zero-resource Hallucination Detection for Text Generation \\
via Graph-based Contextual Knowledge Triples Modeling}
\author {
    Xinyue Fang\textsuperscript{\rm 1},
    Zhen Huang\textsuperscript{\rm 1}\thanks{Corresponding Author},
    Zhiliang Tian\textsuperscript{\rm 1}\footnotemark[1],
    Minghui Fang\textsuperscript{\rm 2},
    Ziyi Pan\textsuperscript{\rm 1},
    Quntian Fang\textsuperscript{\rm 1},
    Zhihua Wen\textsuperscript{\rm 1},
    Hengyue Pan\textsuperscript{\rm 1},
    Dongsheng Li\textsuperscript{\rm 1}
}
\affiliations {
    \textsuperscript{\rm 1}College of Computer, National University of Defense Technology\\
   \textsuperscript{\rm 2}College of Computer Science and Technology, Zhejiang University\\
   \{fangxinyue, huangzhen, tianzhiliang, panziyi, fangquntian, zhwen, hengyuepan, dsli\}@nudt.edu.cn\\
   minghuifang@zju.edu.cn
}

\begin{document}

\maketitle

\begin{abstract}
LLMs obtain remarkable performance but suffer from hallucinations. Most research on detecting hallucination focuses on questions with short and concrete correct answers that are easy to check faithfulness. Hallucination detections for text generation with open-ended answers are more hard. Some researchers use external knowledge to detect hallucinations in generated texts, but external resources for specific scenarios are hard to access. Recent studies on detecting hallucinations in long texts without external resources conduct consistency comparison among multiple sampled outputs. To handle long texts, researchers split long texts into multiple facts and individually compare the consistency of each pair of facts. However, these methods (1) hardly achieve alignment among multiple facts; (2) overlook dependencies between multiple contextual facts. In this paper, we propose a graph-based context-aware (GCA) hallucination detection method for text generations, which aligns facts and considers the dependencies between contextual facts in consistency comparison. Particularly, to align multiple facts, we conduct a triple-oriented response segmentation to extract multiple knowledge triples. To model dependencies among contextual triples (facts), we construct contextual triples into a graph and enhance triples' interactions via message passing and aggregating via RGCN. To avoid the omission of knowledge triples in long texts, we conduct an LLM-based reverse verification by reconstructing the knowledge triples. Experiments show that our model enhances hallucination detection and excels all baselines\footnote{Code:https://github.com/GCA-hallucinationdetection/GCA}. 
\end{abstract}


%

\section{Introduction}

Recent research has shown that large language models (LLMs) achieved state-of-the-art performance in various NLP tasks \cite{fang2024acegenerativecrossmodalretrieval,lu2022sifter}. However, these models often suffer from hallucinations: generate incorrect or fabricated content in a factual way, which undermines models' credibility \cite{ji2023survey,lu2023meta} and limits LLMs' application in fields requiring factual accuracy \cite{huang2023lawyerllamatechnicalreport,su2024hierarchical}. 
Detecting hallucination in the model's responses is crucial for LLMs' boom.

Existing hallucination detection studies mainly concentrate on tasks like question answering (QA) \cite{zhang2023sac3,wen2023grace} and arithmetic calculation \cite{xue2023rcotdetectingrectifyingfactual,tian2022empathetic} with short and concrete correct answers. In these tasks, the consistency among their concrete answers can be easily checked for hallucination \cite{jiang2021can,wang2024intent}. The research on detecting hallucinations in generated long text is more challenging because (1) generating text is open-ended and rarely has concrete answers and (2) long text contains multiple facts and the consistency among them is hard to verify.
Therefore, we focus on hallucination detection in long text generation with black-box models (powerful LLM, like GPT-4) without external resources (i.e. zero-resource setting).

Currently, the studies on black-box zero-resource hallucination detection for text generation can be divided into two categories: 
(1) \textbf{Self-checking} \cite{friel2023chainpollhighefficacymethod,liu2023g} designs prompt texts using chain-of-thought (CoT) to verify response factuality by the LLMs' own capabilities. Though that can be easily implemented and applied in various scenarios, it relies on the model's own ability and can lead to missed detection: The model may overconfidently trust its outputs \cite{li2024think,chen2024insidellmsinternalstates}. (2) \textbf{Consistency comparison} \cite{zhang2023sac3,ma2024largelanguagemodelsunconscious} samples multiple responses to check whether the sampled responses are highly inconsistent, which indicates hallucination \cite{farquhar2024detecting, wang-etal-2024-mitigating}. The method is effective for short responses with few concrete claims, making consistency comparison easy. However, in long text responses, diverse wording or lexical representations of the same fact complicate consistency comparison.

To address those issues, researchers propose \textbf{divide-and-conquer} \cite{zhang2024examinationeffectivenessdivideandconquerprompting} based on consistency comparison. It has three steps: (1) sample multiple additional responses appending to the original response; (2) divide each response into multiple facts; (3) compare the consistency of facts in the original response with those in the sampled responses, where highly inconsistent facts indicate hallucinations. However, it ignores the omission issue: facts in the original but not in sampled responses may be wrongly seen as hallucinations for lack of comparison. To overcome that, \cite{yang2023new,cao2024autohallautomatedhallucinationdataset} propose \textbf{reverse verification}. For each fact in the original response, they prompt LLMs to create a new question answered by the claim and check if the model's response matches the fact. The method can avoid the omission issue because each fact has an accurately matched counterpart for comparison. However, a long text may contain multiple facts $\{A_1, ... A_N\}$, where $A_i$ is the $i$-th fact within LLMs' generation. The above method compares the consistency of each fact sampled from different responses individually (e.g. verify the truthfulness of $A_i$ by comparing with $B_i$, where $B_i$ is $A_i$'s corresponding fact from a sampled response). It ignores the dependencies among each fact that can reduce detection performance. For example, for the response “Einstein won the Nobel Prize for proposing the theory of relativity." we can extract two triples: (Einstein, proposal, theory of relativity) and (Einstein, receive, Nobel Prize). Although each triple is a fact, their dependency is an error. Therefore, considering the dependencies between a fact and its surrounding facts is promising to enhance hallucination detection.

In this paper, we propose a \textbf{graph-based context-aware (GCA)} hallucination detection method for long text generation, which extracts triples to align facts for a better consistency comparison and considers the dependencies between contextual triples. Specifically, to better align facts from responses for consistency comparison, we conduct a triple-oriented response segmentation to extract multiple knowledge triples for each response. Triples are well-structured and thus easy for comparison; in addition, it is easy to align facts among multiple responses since triples carry only facts and get rid of the wording and lexical representations in the response \cite{wang2023structure} (Sec. 3.2). To consider dependencies between a triple and its contextual triples, we construct a graph over a triple and its contextual triples. Then, we encourage triples' interactions via message passing and aggregating features of neighboring nodes via RGCN (relational graph convolutional network) \cite{schlichtkrull2018modeling} (Sec. 3.3). It encourages the interaction between facts as we observe that a certain number of facts has dependencies in a response (See Sec. 4.5). To avoid the omission issue in (Sec. 3.3), we propose an LLM-based reverse verification method with three reconstruction tasks to reconstruct the knowledge triples (Sec. 3.4). These tasks provide a more thorough and detailed detection of each triple in long text responses, enhancing the overall effectiveness of the method. Experiments show that our method effectively improves hallucination detection accuracy for long text responses generated by black-box models under zero-resource conditions.

Our contributions are: (1) we propose a hallucination detection method for long text generation that considers the dependencies between contextual knowledge triples.  (2) We propose a graph-based context-aware hallucination detection via consistency comparison with RGCN. (3) We additionally proposed three reversion verification tasks to help hallucination detection by reconstructing triples. (4) Experiments show that our method outperforms all baselines.

\section{Related Work}
\subsection{White-box Hallucination Detection}
These methods analyze the model's internal states to identify hallucinations \cite{yuksekgonul2024attentionsatisfiesconstraintsatisfactionlens,lu2023damstf,wen2024perceptionknowledgeboundarylarge}, mainly divided into two types:
(1) \textit{Output logit based method:} The model's output logits reflect the confidence of the model's predictions \cite{jiang2024large}. \cite{varshney2023stitchtimesavesnine} calculates the logits for concepts in response and takes the minimal probabilities to model the uncertainty. \cite{verma2023reducingllmhallucinationsusing} explores the integration of Epistemic Neural Networks (ENNs) with LLMs to improve the model's uncertainty estimates. \cite{luo2023zeroresourcehallucinationpreventionlarge} proposes to adjust the model's output logits by adding a linear layer to better align with correctness likelihood.
(2) \textit{Hidden layer activations based method:} Hidden layer activations encapsulate the model's internal representation of statement truthfulness \cite{fadeeva2024factcheckingoutputlargelanguage}. \cite{azaria2023internal} trains a classifier using LLM's hidden layer activations to assess the truthfulness of statements. \cite{snyder2024early} uses output values from artifacts associated with model generation as input features to train a classifier that identifies hallucinations. \cite{zhu2024pollmgraph} uses probabilistic models to analyze internal state transitions in the LLM during generation to detect hallucinations.

\subsection{Black-box Hallucination Detection using External Resources}
These methods aim to verify the authenticity of model-generated content by leveraging external knowledge \cite{wen2023grove,nahar2024fakesvaryingshadeswarning}. Depending on the source of external knowledge, it can be categorized into the following two types.
(1) \textit{RAG-based method:} Retrieval-augmented generation (RAG) is a technique that enhances text generation by retrieving relevant information from external sources \cite{sadat2023delucionqa,wang2023canonicalization}. \cite{roychowdhury2024erattaextremeragtable} proposes a multi-LLM system with the capability to retrieve external knowledge bases and perform real-time content authentication. \cite{ding2024retrieveneedsadaptiveretrieval} retrieves relevant evidence to help LLMs correct potential hallucinations in responses. \cite{kang2024evermitigatinghallucinationlarge} proposes a method for the real-time retrieval of Web search engines that can verify the factuality of the output responses and correct hallucinations. \cite{li2024drowzeemetamorphictestingfactconflicting} automatically retrieves knowledge graphs to detect hallucinations through logical programming and mutation testing. Furthermore, \cite{bayat2023fleek} proposes an automated method to extract factual claims from responses and collect evidence from knowledge graphs to verify the factuality of the claims to be extracted.
(2) \textit{Incorporating Alternative Models:} Researchers use responses generated by other models for cross-validation to detect hallucinations \cite{hegselmann2024datacentricapproachgeneratefaithful}. \cite{cohen2023lm} constructs a framework for assessing the factuality of output responses through cross-validation by two language models. \cite{rawte2024factoidfactualentailmenthallucination,wan2024dellgeneratingreactionsexplanations} use multiple LLMs as ``judges'' to evaluate various aspects of the model's output responses. \cite{li2024drowzeemetamorphictestingfactconflicting} proposes a mutation testing model based on logic programming, which can verify the consistency of LLMs' responses with real-world situations.

\subsection{Black-box Hallucination Detection using Zero-resource} Researchers propose using the model's own capabilities to detect hallucinations \cite{liu2024enablingweakllmsjudge} because obtaining high-quality external resources is challenging \cite{mündler2024selfcontradictoryhallucinationslargelanguage}.
(1) \textit{For non-long text responses generated by the model:} Consistency comparison through multiple sampling responses is an important method \cite{allen2024shroomindelabsemeval2024task6}. \cite{zhang2023sac3} improves hallucination detection performance on commonsense QA tasks through semantic-aware cross-validation consistency. \cite{liu2024enablingweakllmsjudge} evaluates the reliability of responses generated by LLMs for individual questions or queries through cross-query comparison. \cite{ma2024largelanguagemodelsunconscious} proposes a critical calculation and conclusion (CCC) prompt template to enhance LLM's ability to detect and correct unreasonable errors in mathematical problem-solving. \cite{yehuda2024interrogatellmzeroresourcehallucinationdetection} identifies potential instances of hallucination by quantifying the level of inconsistency between the original query and the reconstructed query.
(2) \textit{For long text responses generated by the model:} \cite{manakul2023selfcheckgpt} proposes a method to detect hallucinations by comparing the consistency of responses from multiple random samplings. \cite{yang2023new} introduces a reverse validation method for passage-level hallucination detection in LLMs, which leverages the LLM's own knowledge base. \cite{mündler2024selfcontradictoryhallucinationslargelanguage} introduces a method for detecting self-contradictions in long text responses through logical reasoning. \cite{friel2023chainpollhighefficacymethod} proposes an efficient prompting method that uses the chains of thought generated by LLMs to detect hallucinations in the responses. Unlike these LLM-based methods, our approach constructs long text responses as graphs and uses graph neural networks to capture the contextual influence of each fact during hallucination detection.
\begin{figure*}[htp]
    \centering
    \includegraphics[height=9cm, width=17.5cm]{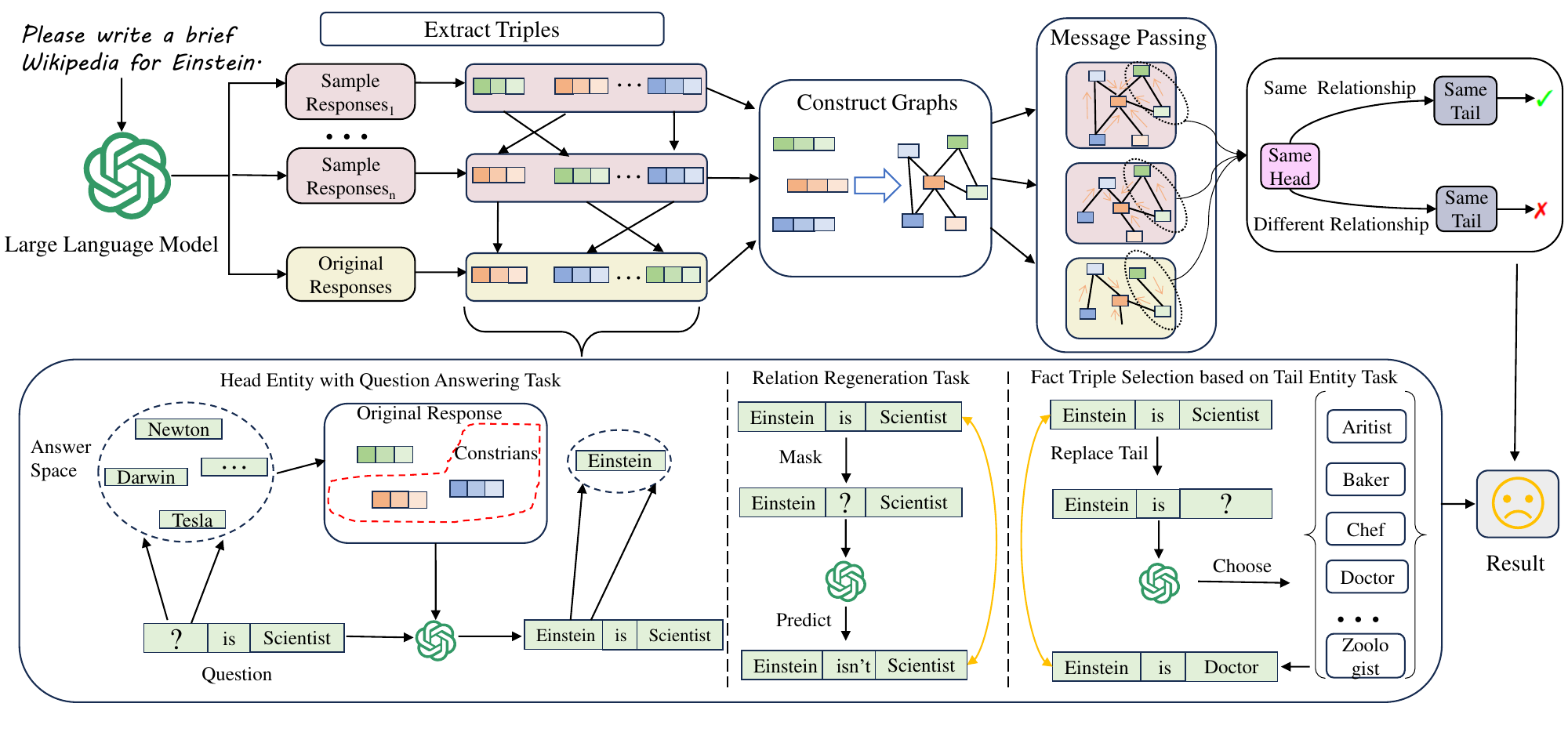}
    \caption{GCA framework. We extract triples from the original response and sampled responses (left-upper corner). Then, we construct a graph for each response with the extracted triples and perform message passing and aggregation on the graph (as the upper branch). We conduct reverse validation for each part of the triples with three reconstruction tasks (as the lower branch).}
    \label{fig:framework}
\end{figure*}
\section{Method}
\subsection{Overview}
Our method has three modules: (1) \textbf{Triple-Oriented Response Segmentation} (Sec. 3.2) extracts facts from the model's responses. (2) \textbf{Graph-based Contextual Consistency Comparison with RGCN} (Sec. 3.3) constructs a graph carrying the extracted knowledge triples and utilizes an RGCN to propagate and integrate messages across the graph. It considers the dependencies between each knowledge triple (facts) and its surrounding triples during detection. (3) \textbf{Reverse Verification via Triples Reconstruction} (Sec. 3.4) achieves reverse verification for hallucination detection by reconstructing each triple via three LLM-based tasks (as shown in Fig.\ref{fig:framework}).

We feed each knowledge triple extracted (Sec. 3.2) to detect hallucinations (Sec. 3.3 and Sec. 3.4), and then we judge the original long text response relying on the results of each triple from (Sec. 3.3 and Sec. 3.4).

\subsection{Triple-Oriented Response Segmentation}\label{sec:3.2}
To better align facts in the consistency comparison, we propose to segment the responses by extracting knowledge triples as facts and checking the answers' consistency among the triples. Our motivation is that due to the impact of wording, comparing textual consistency can lead to mismatches. Because hallucination detection considers the semantics of knowledge fact instead of specific word choices, we use a triple-based comparison method to provide better alignment than traditional textual comparison. Specifically, the steps are as follows:

\begin{itemize}
\item \textbf{Extraction.} Inspired by the latest method \cite{hu2024refcheckerreferencebasedfinegrainedhallucination}, we design prompts to extract knowledge triples from responses using an LLM.
\item \textbf{Verification.} To ensure the accuracy of the extracted knowledge triples, we pair each response with its triples and prompt the LLM to confirm their semantic equivalence. If any ambiguities exist between the extracted triples and the response text, we instruct the LLM to adjust the semantics of the triples according to the text's semantics. The details of the prompts are in App.A.
\end{itemize}

Knowledge triples have a structured format and are easy to compare, simplifying alignment and comparing consistency between responses, enhancing detection accuracy.

\subsection{Graph-based Contextual Consistency Comparison with RGCN}\label{sec:3.3}
To effectively consider dependencies between triples, we propose Graph-based Contextual Consistency Comparison \textbf{(GCCC)}, which constructs a knowledge graph for each response and then conducts message passing and aggregation via RGCN. The intuition is that traditional consistency comparison focuses on comparing individual facts: it verifies a single piece of knowledge fact by comparing it only with the corresponding fact in the sampled responses at a time. It results in ignoring the triples that are mutually dependent on the given triple within the context information. 

To address the problem, our approach constructs graphs for the original response and the sampled responses. Then, it employs RGCN for message passing and aggregation on these graphs. The process consists of two stages: (1) \textit{knowledge triples modeling via graph learning}. We build a graph for each response and obtain node (entity) embeddings via RGCN processing to model the multiple knowledge triples for a response. (2) \textit{triples consistency comparison}. We compare the consistency of triples across the graphs at the embedding level.

\subsubsection{Knowledge Triples Modeling via Graph Learning}
This stage is divided into three steps: firstly, we convert each response into a graph. Then, we obtain the initial node (entity) embeddings for each graph using sentence-BERT \cite{wang2020minilm}. Finally, we employ the RGCN to perform message passing and aggregation using the initial node embeddings on each graph, updating the node embeddings. 
\begin{itemize}
\item \textbf{Graph Construction.} For a user's query, the model generates an original response \( R_{\text{o}} \), and we sample multiple additional responses \( R_{\text{sampled}} = \{R_{1}, R_{2}, \ldots, R_{n}\} \). \((h_i, r_i, t_i)\) is a single triple in \( R_{\text{o}} \) and \( (h_{i,j}, r_{i,j}, t_{i,j})\) is a single triple in \( j\)-th sampled response \( R_{j} \).  We construct the graph \( G_{\text{o}} = (V_{\text{o}}, E_{\text{o}}) \) for the original response, in which vertices \((v \in V_{\text{o}}) \) represent the head and tail entities from each triple. An edge \((e\in E_{\text{o}}) \) represents the relation between the head entity and the tail entity. Similarly, we construct the graph \( G_{j} = (V_{j}, E_{j}) \) for each sampled response \( R_{j} \). By doing so, we construct several response-specific graphs for each user's query.
\item \textbf{Representation Initialization.} Using sentence-BERT, we encode the head entities, relation, and tail entities in knowledge triples as vector representations. For the original response, we represent each triple embedding as: \( (\mathbf{h}_{i}, \mathbf{r}_{i}, \mathbf{t}_{i}) = \text{BERT}(h_i, r_i, t_i) \). For each sampled response, we represent each triple embedding as: \( (\mathbf{h}_{i,j}, \mathbf{r}_{i,j}, \mathbf{t}_{i,j}) = \text{BERT}(h_{i,j}, r_{i,j}, t_{i,j}) \). We treat the head and tail entity embeddings from the original response as \( G_{\text{o}} \)'s initial node (entity) embeddings. Similarly, we obtain the initial node (entity) embeddings for the graph \( G_{j} \) corresponding to  \( j \)-th sampled response.
\item \textbf{Message Passing and Aggregation.}
We use the RGCN to perform message passing and aggregation on the graph. As Eq.\ref{eq:3} shows that for each layer \( l \), the new representation of each node \( v \) is denoted as \(\mathbf{e}_v^{(l+1)} \). For each relation \(r \in \mathcal{R}\), we denote the set of all neighbors of the node \( v \) that are connected through an edge of relation \( r\) as \(\mathcal{N}_r(v)\). For each neighbor in \(\mathcal{N}_r(v)\), we multiply its representation \(\mathbf{e}_u^{(l)} \) by a weight matrix \( \mathbf{W}_r^{(l)} \) and normalize it using the hyperparameter \(c_{v,r}\). In addition to aggregating information from neighbors, \(\mathbf{e}_v^{(l+1)} \) also includes its own representation \( \mathbf{e}_v^{(l)} \) from the previous layer \(l\) and transform it by a weight matrix \( \mathbf{W}_0^{(l)}\).
\begin{small}
\begin{equation}{\label{eq:3}}
\mathbf{e}_v^{(l+1)} = \sigma \left( \sum_{r \in \mathcal{R}} \sum_{u \in \mathcal{N}_r(v)} \frac{1}{c_{v,r}} \mathbf{W}_r^{(l)} \mathbf{e}_u^{(l)} + \mathbf{W}_0^{(l)} \mathbf{e}_v^{(l)} \right)
\end{equation}
\end{small}
\end{itemize}
The updating for \(\mathbf{e}_v^{(l+1)} \) integrates information from \( v \)'s neighbors through relation-specific change, while also incorporating \( v \)'s own representation. These operations ensure that the updated node embedding is informed by its context and intrinsic properties. Triples containing the node can also incorporate contextual information, enhancing the accuracy when comparing the consistency of triples, thereby improving the detection of hallucinations.
 
\subsubsection{Triples Consistency Comparison.}
Based on the graph representations from RGCN, we detect hallucinations in the original response by comparing the consistency of triples across multiple graphs. Firstly, we align triples between the original response's graph \( G_{o} \) and each sampled graph \( G_{j} \). Then we compare the consistency of the aligned triples to calculate the consistency score. 
\begin{itemize}
\item \textbf{Triples Alignment.}
For each triple \((h_i, r_i, t_i)\) in the original response and each triple \( (h_{i,j}, r_{i,j}, t_{i,j})\) in the sampled response, we first check whether the head entities of these two triples are the same. If so, we calculate the similarity \({S}(\mathbf{r}_{i}, \mathbf{r}_{i,j})\) between the relation representation \( \mathbf{r}_{i}\) of relation \( r_i \) and the representation \( \mathbf{r}_{i,j}\) of \( r_{i,j} \). If \({S}(\mathbf{r}_{i}, \mathbf{r}_{i,j})\) exceeds the pre-defined threshold \( \theta_r \), we regard the two triples as aligned. Otherwise, they are considered unaligned. For every triple in the original response, we apply the above operations to align each triple from sampled responses with it. 
\item \textbf{Consistency score calculation.}
After aligning the triples, we need to further compare whether they are consistent with each other to calculate the consistency score. Specifically, as    Eq. \ref{eq:4} shows, for a triple \((h_i, r_i, t_i)\) in the original response and its aligned triple \( (h_{i,j}, r_{i,j}, t_{i,j})_{\text{a}}\) in \( j \)-th sampled response, \(\mathbf{e}_i\) and \(\mathbf{e}_{i,j}^{\text{a}}\) are the node embeddings of the tail entity \({t}_i\) and \( t_{i,j} \) after RGCN processing. We compute the similarity between \(\mathbf{e}_i\) and \(\mathbf{e}_{i,j}^{\text{a}}\). If their similarity \({S}(\mathbf{e}_{i}, \mathbf{e}_{i,j}^{\text{a}})\) exceeds the threshold \( \theta_t\), we increase the consistency score \( c_{i,j} \) of \((h_i, r_i, t_i)\) by 1. This indicates that there is a triple consistent with the triple \((h_i, r_i, t_i)\) in \(j\)-th sampled response. 

Conversely, we use \(\mathbf{e}_{t_{i,j}}^{\text{m}}\) to denote the node embedding of the tail entity in the unaligned triple \( (h_{i,j}, r_{i,j}, t_{i,j})_{\text{m}}\) in the \( j \)-th sampled response. If the similarity between \(\mathbf{e}_{t_{i,j}}^{\text{m}}\) and \(\mathbf{e}_i\) exceeds the threshold \( \theta_t\), we update the consistency score \( c_{i,j} \) of \((h_i, r_i, t_i)\) by subtracting 1. It indicates that the triple may have a risk of hallucination. Note we do not directly label the triple as a hallucination, as two triples with the same head and tail entities but different relations can both be factually correct. Moreover, such cases are rare (1.9\% in two datasets for hallucination detection), as the knowledge triples we compare for consistency come from repeated responses to the same query, which are likely to focus on describing the same subject matter. In Sec. 3.4, we also provide a detailed detection for each triple to ensure the accuracy of the results.
\begin{equation}\label{eq:4}
c_{i,j} = 
\begin{cases} 
c_{i,j} + 1 & \text{if } \text{S}(\mathbf{e}_{t_i}, \mathbf{e}_{t_{i,j}}^{\text{a}}) > \theta_t \\
c_{i,j} - 1 & \text{if } \text{S}(\mathbf{e}_{t_i}, \mathbf{e}_{t_{i,j}}^{\text{m}}) > \theta_t \\
c_{i,j} & \text{otherwise}
\end{cases}
\end{equation}
To obtain the final consistency score for each triple in the response, we sum its comparison results with each sampled response as 
$C_{i} = \sum_{j=1}^{n} \left( c_{i,j} \right)$
\end{itemize}

During the message passing and aggregation process with RGCN on a graph, each node integrates features from its neighboring nodes. This allows triples containing the node to aggregate contextual information from surrounding triples, Considering the dependencies between the triple to be verified and the surrounding triples.

\subsection{Reverse Verification via Triple Reconstruction}\label{sec:3.4}
To address the omission issue mentioned in Sec. 3.3, we propose a LLM-based reverse verification method (\textbf{RVF}), which contains three reconstruction tasks that check whether LLM can reconstruct the knowledge triples' head entity, relation, and tail entity, respectively. Traditional reverse strategies prompt the LLMs to reconstruct questions to verify each knowledge fact from generated responses. The reconstructed question may have multiple correct answers, which leads to a low probability of answering the facts that we aim to verify. It increases the chance of misjudging these facts. To address this, we add constraints to the reconstructed questions to reduce the space of correct answers and increase the probability of answering the triples we want to verify. The three tasks are as follows: 
\begin{itemize}
     \item \textbf{Head Entity with Question Answering Task (HEQA).} We prompt LLMs to reconstruct a question for each triple, with the head entity as the expected answer, and then obtain the model's responses. We check if these responses are consistent with the head entity. Specifically, to reduce the space of correct answers for reconstructed questions, we first follow the method from \cite{manakul2023selfcheckgpt} to initially verify the triples in the original responses. Then, we filter out a set of triples \( f_{t}\) with high factual accuracy. For each triple \( (h_{i}, r_{i}, t_{i}) \) in the original response, we add \( f_{t} \) (excluding \( (h_{i}, r_{i}, t_{i}) \) if it exists in \( f_{t}\)) as constraints in the questions during the LLM reconstruction process. The model's responses to the question must satisfy these constraints. We repeatedly prompt the LLM to generate answers \(A\) to the question. The total number of \(A\) denoted as \( N_{A} \). We count the times that the model responses match the head entity \( h_{i} \) (denoted as \( N_{h} \)) and calculate the fact score \(S_{h}\) as the ratio of \( N_{h} \) to \( N_{A} \), where \( S_{h} = \frac{N_{h}}{N_{A}}\).
    \item \textbf{Relation Regeneration Task (RR).} We mask the relation in the triple with a special token and prompt the model to predict multiple times. Then we check whether the model's predictions are identical to the relation for measuring the consistency. It can reduce the space of correct answers because the relationship between two entities is limited. Specifically, for each triple \( (h_{i}, r_{i}, t_{i}) \), we mask \( r_{i} \) with a special token and prompt the LLM for multiple times to predict the original \( r_{i} \) given \( h_{i} \) and \( t_{i} \). We define the fact score \( S_{r} \) as the proportion of the predicted relations that match the original relation \( r_{i} \),  where \(S_{r} = \frac{N_{c}}{N_{p}}\). Here, \( N_{c} \) is the number of matched predictions, and \( N_{p} \) is the total number of predictions.
     \item \textbf{Fact Triple Selection based on Tail Entity Task (FTSTE).} Models often generate long texts centered around a few key entities, which typically serve as the head entities in extracted triples. The limited number of head entities allows us to use surrounding context related to the head entity as constraints to effectively narrow down the space of correct answers for reconstructed questions. However, tail entities in long-text responses are more diverse, so we cannot directly use surrounding contexts as constraints in reconstructed questions. Instead, we use a direct approach by providing a list of options to limit the space of correct answers. We prompt the model to select the factual triple from it.  Then, we compare if the model’s selections are consistent with the original triple. It reduces the space of correct answers by providing a limited set of choices. Specifically, for each triple \( (h_{i}, r_{i}, t_{i}) \), we replace \(  t_{i} \) with other entities of the same type to generate multiple similar triplets; and then, we prompt the LLM to choose the factual one. We define the fact score \( S_{t} \) as the proportion of times \( (h_{i}, r_{i}, t_{i}) \) is selected, where \(S_{t} = \frac{N_{t}}{N_{s}}\). \(N_{t}\) is the number of times \( (h_{i}, r_{i}, t_{i}) \) is selected, and \(N_{s}\) is the total number of selections. See the prompt templates used in the above three tasks in App.B.
\end{itemize}

Finally, we sum up the fact scores from these three tasks and the consistency score mentioned in Sec. 3.3 with different weights to make a judgment about each triple in the original response, as shown in Eq. \ref{eq:6}
\begin{equation}{\label{eq:6}}
   F(h_i, r_i, t_i) = w_1 \cdot S_{h} + w_2 \cdot S_{r} + w_3 \cdot S_{t} + w_4 \cdot C_{i}
\end{equation}

In our proposed reverse detection method, the three tasks use different strategies to reduce the space of correct answers in the reconstructed questions. It avoids the issue in traditional reverse detection techniques where the reconstructed questions may have multiple correct answers making it difficult to detect specific facts, improving the accuracy of detecting each triple in the original response.
\section{Experiments}
\subsection{Experimental Setting}
\subsubsection{Datasets}
We utilize three datasets: (1) \textbf{PHD} \cite{yang2023new}: The dataset consists of 300 samples. Each sample is a Wikipedia article about an entity generated by ChatGPT (gpt-3.5-turbo) and annotated by human annotators. (2) \textbf{WikiBio} \cite{manakul2023selfcheckgpt}: The dataset consists of 238 passages generated by GPT3 and annotated at the sentence level. Although it lacks passage-level labels, we follow \cite{yang2023new} to aggregate sentence labels to derive pseudo-labels at the passage level. (3) \textbf{sub-WikiBio}: There are only 12 fact samples in the WikiBio dataset. The sample distribution is too imbalanced. Therefore, we extract all 12 fact samples and 48 randomly selected hallucination samples to create a subset. In our experiment, we refer to it as the WikiBio subset. 
\subsubsection{Implemention Details}
We use the recorded responses for each sample as original responses and generate 10 additional sampled responses using ChatGPT. we set the generation temperature to 1.0 to ensure the randomness of sampled responses. We use GPT-4 (gpt-4-1106-preview) to extract triple knowledge from responses and reconstruct questions in reverse verification. At this point, we set the temperature to 0.0 to maximize the reproducibility of the result.
\subsubsection{Baselines}
We compare our method against six baselines: (1) \textbf{Reverse Validation via QG (RVQG)} \cite{yang2023new} is a method that uses LLMs to reconstruct a question about the text to be verified. It compares if the model's response to the reconstructed question is consistent with the text (2) \textbf{Semantic Entropy (SE)} \cite{farquhar2024detecting} breaks down the entire response into factual claims and prompts LLMs to reconstruct questions about it. For each claim, they repeatedly ask LLM reconstructed questions. And then cluster the claim and the model's responses. They measure the entropy of the cluster containing the claim to assess its validity. (3) \textbf{SelfCheckGPT via BERTScore (SelfCk-BS)} \cite{manakul2023selfcheckgpt} is a variant of SelfCheckGPT, using BERTScore to measure consistency between original response and sampled responses. (4) \textbf{SelfCheckGPT via NLI (SelfCk-NLI)} \cite{manakul2023selfcheckgpt} is another variant of SelfCheckGPT that uses an NLI model to measure consistency between the original response and the sampled responses. (5) \textbf{Self-contradiction (SC)} \cite{mündler2024selfcontradictoryhallucinationslargelanguage} is a prompting-based framework designed to effectively detect self-contradictory hallucinations. (6) \textbf{Focus} \cite{zhang2023enhancing}  is a white-box hallucination detection method that works by focusing on the properties of key tokens in the response.
However, SE, SelfCk-BS, SelfCk-NLI, and Focus all return the likelihood scores of a sample being a hallucination, rather than labels indicating fact or hallucination. To align these methods with our task,  we set thresholds for these baselines on different datasets using the same approach as for our method. If a sample score exceeds the threshold, we classify it as a hallucination. Details are in App.C.
\subsubsection{Evaluation Metrics.} We evaluate how well the method detects hallucinatory responses using metrics: (1) \textbf{F1} is the harmonic mean of precision and recall, providing a comprehensive evaluation of the classification performance of the method; (2) \textbf{Accuracy} is the proportion of correctly classified samples out of the total number of samples.
\begin{table}[htb]
\centering
\scalebox{0.8}{
\begin{tabular}{c|cc|cc|cc}
\hline
\multirow{2}{*}{\textbf{\centering Methods}} & \multicolumn{2}{c|}{\textbf{PHD}} & \multicolumn{2}{c|}{\textbf{WikiBio}} & \multicolumn{2}{c}{\textbf{sub-WikiBio}} \\
                                  & \textbf{F1}          & \textbf{Acc}         & \textbf{F1}          & \textbf{Acc}        & \textbf{F1}            & \textbf{Acc}          \\ \hline
RVQG         & 52.3                 & 65.3                 & 85.7                 & 79.2                & 88.2                   & 81.7                  \\
SE                  & 35.6                 & 62.7                 & 66.7                 & 52.5                & 87.9                   & 82.3                  \\
SelfCk-NLI                     & 23.7                 & 42.0                 & 60.2                 & 43.3                & 44.7                   & 30.0                  \\
SelfCk-BS                   & 40.5                 & 55.0                 & 71.0                 & 57.1                & 88.8                   & 83.3                  \\
SC                     & 30.9                 & 65.7                 & 75.8                 & 62.2                & 83.7                   & 76.7                  \\
Focus                   & 46.7                 & 62.0                 & 75.7                 & 61.3                & 83.3                   & 76.7                  \\
GCA                          & \textbf{55.4}        & \textbf{68.3}        & \textbf{90.7}        & \textbf{83.2}       & \textbf{90.5}          & \textbf{85.0}         \\ \hline
\end{tabular}
}
\caption{Evaluation results of all methods.}
\label{tb:1}
\end{table}

\subsection{Overall Performance}
We analyze the effectiveness of our method by comparing it with six baselines, results shown in Tab.\ref{tb:1}. Our method outperforms baselines on all metric values. SelfCk-NLI uses an NLI model to assess if any sentence in the original response contradicts sampled responses and performs the worst on all metrics. SelfCk-NLI does not perform as well as SelfCk-BS, suggesting that NLI models have limited ability to compare consistency between texts. It is even less effective than assessing via simple embedding similarity measures. Reverse validation methods (RVQG and SE) perform worse than our method on all metrics. We attribute this to using a graph-based consistency comparison method (Sec. 3.3), which considers dependencies between triples during comparison. Notably, our method outperforms Focus, a white-box hallucination detection method that uses internal model information, further demonstrating its outstanding efficacy.
\begin{table}[htb]
\centering
\scalebox{0.8}{
\begin{tabular}{c|cc|cc|cc}
\hline
\multirow{2}{*}{\textbf{\centering Variants}} & \multicolumn{2}{c|}{\textbf{PHD}} & \multicolumn{2}{c|}{\textbf{WikiBio}} & \multicolumn{2}{c}{\textbf{sub-WikiBio}} \\
\multicolumn{1}{c|}{}                                   & \textbf{F1}          & \textbf{Acc}         & \textbf{F1}          & \textbf{Acc}        & \textbf{F1}            & \textbf{Acc}          \\ \hline
$-$ RVF & 38.1                 & 40.3                 & \textbf{90.7}        & \textbf{83.2}       & 88.7                   & 80.0                  \\
$-$ GCCC                   & 54.0                 & 67.7                 & 87.1                 & 78.2                & 87.6                   & 81.7                  \\
$-$ RR                                   & 52.1                 & 65.6                 & 87.0                   & 77.7                & 90.0                   & 83.3                  \\
$-$ FTSTE                                  & 52.1                 & 66.3                 & 86.8                 & 77.7                & 89.1                   & 83.3                  \\
$-$ HEQA                                      & 36.4                 & 54.6                 & 84.5                 & 73.9                & 85.1                   & 75.0                  \\
$-$ Relations                                                 & 53.7                 & 66.7                 & 86.8                 & 77.7                & 88.6                   & 80.0 \\
$-$ Graph                                               & 52.8                 & 66.7                 & 83.7                 & 73.1                & 87.6                   & 81.7                  \\ \hline
\multicolumn{1}{c|}{GCA}                           & \textbf{55.4}        & \textbf{68.3}        & \textbf{90.7}        & \textbf{83.2}       & \textbf{90.5}          & \textbf{85.0}         \\ \hline
\end{tabular}
}
\caption{Ablation studies on model components. $-$RVF and $-$GCCC respectively means detecting without RVF and GCCC. $-$ RR, $-$ FTSTE and $-$ HEQA respectively indicate removing the RR task, FTSTE task, and HEQA task from the full model. $-$ Relations means detecting without relations in triples. $-$ Graph means detecting without graph network model.}
\label{tb:2}
\end{table}
\subsection{Ablation study}
We conduct an ablation study to verify the importance of each component as shown in Tab \ref{tb:2}. \textit{$-$ RVF} means abandoning the reverse validation from our full model. The performance drop across most datasets indicates that RVF effectively addresses the omission issues in GCCC to improve the overall effectiveness. However, the performance does not drop on the WikiBio. The reason is that WikiBio contains many hallucination samples (95\%), causing our method, baselines, and their variants to show a bias toward predicting hallucinations in this dataset. In these abnormal conditions, the RVF module does not perform effectively, as its advantage lies in correctly identifying hallucination samples. With a more balanced sample distribution in the dataset (sub-WikiBio), our full model performs better than GCCC as expected. \textit{$-$ GCCC} removes GCCC from the full model, performing worse than GCA. It indicates that GCA utilizes GCCC to consider the dependencies between triples in the consistency comparison process, improving the accuracy of results.\textit{$-$ RR}, \textit{$-$ FTSTE} and \textit{$-$ HEQA} respectively represent removing the RR task, FTSTE task, and HEQA task mentioned in Sec.3.4 from our full model. $-$ HEQA shows the worst performance, indicating that the HEQA task is the most effective reverse detection strategy. \textit{$-$ Relations} means not utilizing the relations in the triples during the consistency comparison process. It replaces the RGCN used in GCCC with GCN, and the results show a decline. It suggests that relation information is useful and RGCN effectively integrates it for each triple. \textit{$-$ Graph} means not using graph network, performing worse than GCA. It indicates that information integrated by RGCN is beneficial for detection.

\subsection{Analysis on Contextual Integration in GCCC}
To verify that our graph-based method effectively aggregates node information, we design an experiment to compare two scenarios: \textbf{(S1)} using RGCN for message passing and aggregation on the graph; \textbf{(S2)} without RGCN, examining the similarity between nodes and their surrounding neighbors. Specifically, we conduct two experiments as follows: 
\begin{figure}[htp]
    \centering
    \includegraphics[height=7.7cm, width=8.4cm]{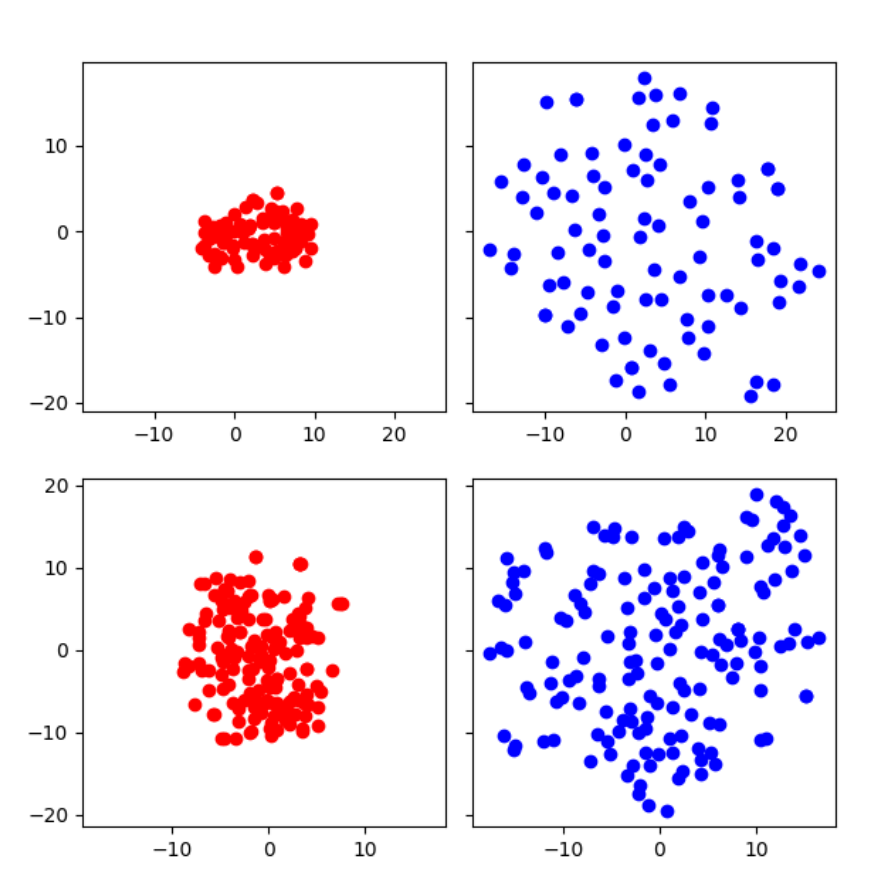}
    \caption{Node distribution comparisons with (red points) and without (blue points) RGCN on the PHD (top) and WikiBio (bottom).}
    \label{fig:Nodes distribution}
\end{figure}

\paragraph{\textbf{t-SNE Visualization of Node Representation Distribution.}} The first experiment uses t-SNE dimensionality reduction to project the node representations from both scenarios into a two-dimensional space to observe the distribution. Fig.\ref{fig:Nodes distribution} shows that in both the PHD and the WikiBio, the node distribution in the \textbf{(S1)} (red nodes) is more compact compared to the \textbf{(S2)} (blue nodes). This indicates that after using RGCN, the node representations become more similar to those of their neighbors. RGCN effectively integrates the features of neighboring nodes into each node's representation, blending information for every node.

\paragraph{\textbf{Quantitative Analysis of Node Representation Similarity.}} We perform a quantitative analysis by obtaining the cosine similarity of node representations under both \textbf{(S1)} and \textbf{(S2)}. Tab.\ref{tb:analyze1} shows that the representations' similarity between two nodes is higher after processing with RGCN compared to without RGCN. This also indicates that our method integrates contextual information for each node by using RGCN.
\begin{table}[htb]
\centering
\scalebox{0.8}{
\begin{tabular}{c|cc|cc}
\hline
\multirow{2}{*}{\textbf{\centering Methods}} & \multicolumn{2}{c|}{\textbf{PHD}}               & \multicolumn{2}{c}{\textbf{WikiBio}}                      \\
                                  & \textbf{Min} & \textbf{Avg} & \textbf{Min} & \textbf{Avg} \\ \hline
GCCC                              & 0.277        & 0.486        & 0.401        & 0.557        \\
$-$ RGCN                           & 0.189        & 0.305        & 0.167        & 0.247        \\ \hline
\end{tabular}
}
\caption{Similarity of representations between nodes. GCCC means our full graph-based module GCCC. $-$ RGCN indicates not using RGCN in the GCCC module. App. D shows the full version of the results.}
\label{tb:analyze1}
\end{table}

\subsection{Triple Dependencies Error Detection}
 We design an experiment to verify that our method can also detect errors in triple dependencies. Specifically, we create a new dataset, TripleCom, by selecting samples with errors in triple dependencies extracted from the PHD and WikiBio datasets. The proportion of such error is approximately 10.5\% in these datasets. Then we test GCA and six baselines on this dataset, with implementation details matching those in Sec.4.1.2. As shown in Tab. \ref{tb:analyze2}, our method GCA achieves the best performance on all metrics in the TripleCom dataset,  demonstrating its effectiveness in detecting errors in the dependencies between multiple triples.

\begin{table}[htb]
\centering
\scalebox{0.8}{
\begin{tabular}{c|cc}
\hline
\textbf{Methods} & \textbf{F1} & \textbf{Acc} \\ \hline  
RVQG             & 81.1       & 68.2      \\
SE               & 56.3       & 39.1      \\          
SelfCk-BS        & 66.7       & 50.0      \\
SelfCk-NLI       & 66.7       & 50.0      \\
SC        & 70.6       & 54.5      \\
Focus       &75.7       & 60.1     \\
GCA         & \textbf{92.7} & \textbf{86.3} \\ \hline
\end{tabular}
}
\caption{Results of methods on the TripleCom dataset.}
\label{tb:analyze2}
\end{table}

\section{Conclusion}
In this paper, we propose a graph-based context-aware hallucination detection method on long-text generation, where our method follows a zero-resource setting and uses only black-box LLMs. Our method extracts knowledge triples from output responses for better alignment. We then construct a graph to carry contextual information so that considers dependencies between knowledge triples. It indeed addresses the issue of ignoring the contextual information in existing methods that only focus on individual facts. We construct three reconstruction tasks for reverse verification to verify the knowledge triples. Experiments show that our method outperforms all baselines, including the white-box method with access to internal model information, excelling in hallucination detection. 
\section*{Acknowledgements}
This work is supported by the following fundings:  National Natural Science Foundation of China under Grant No. 62376284 and No. 62306330, Young Elite Scientist Sponsorship Program by CAST (2023QNRC001) under Grant No. YESS20230367.

\bibliography{aaai25}

\clearpage
\appendix
\section{A Details of Triple Extraction and Verification Prompt Templates}
Table \ref{tb:1} shows the prompts for extracting and verifying triplets mentioned in Sec. 3.2. For extracting triplets, the prompt uses a few-shot setup by adding two examples in the context. In contrast, the prompt for verifying the extracted triplets does not include any examples.
\begin{table*}[ht]\label{tb:1}
    \centering
    \scalebox{0.9}{
    \begin{tabularx}{\textwidth}{>{\centering\arraybackslash}c | X}
        \hline
        \textbf{\centering Prompt Type} & \textbf{\centering Prompt Templates} \\ \hline
        \textbf{\centering Extraction} & In the knowledge graph, knowledge triples are a basic data structure used to represent and store information, and each triple is an expression of a fact. Given a piece of text, please extract all knowledge triples contained in the text, and represent the triples in the form of ("head entity", "relationship", "tail entity"). \newline
        Note that the extracted triples need to be as fine-grained as possible. It is necessary to ensure that the semantics of the triple are consistent with the information in the corresponding part of the text and there is no pronoun in the triple.  All knowledge triples in the text need to be extracted. \newline
        Here is an in-context example: \newline
        \texttt{<Response>}: Paris, the capital of France, is a city with a long history and full of romance. Not only is there the world-famous Eiffel Tower and Louvre Museum, but it also has a unique artistic atmosphere and rich cultural heritage. \newline
        \texttt{<Triples>}:
        Triple: (Paris, is, the capital of France)
        Triple: (Paris, possession, long history)
        Triple: (Paris, full, romantic)
        Triple: (Paris, possession, Eiffel Tower)
        Triple: (Paris, possession, Louvre)
        Triple: (Paris, possessions, unique artistic atmosphere)
        Triple: (Paris, possessions, rich cultural heritage) \newline
        \texttt{<Response>}:"\"The Girl Who Loved Tom Gordon\" is a novel by Stephen King, published in 1999. The story follows a young girl named Trisha McFarland who becomes lost in the woods while on a family hike. As she struggles to survive, she turns to her favorite baseball player, Tom Gordon, for comfort and guidance. The novel explores themes of isolation, fear, and the power of imagination. It was a critical and commercial success and has been adapted into a comic book and a stage play. \newline
        \texttt{<Triples>}:
        Triple: ("The Girl Who Loved Tom Gordon", is, a novel by Stephen King)
        Triple: ("The Girl Who Loved Tom Gordon", published in, 1999)
        Triple: ("The Girl Who Loved Tom Gordon", follows, Trisha McFarland)
        Triple: ("The Girl Who Loved Tom Gordon" protagonist: Trisha McFarland, becomes, lost in the woods)
        Triple: ("The Girl Who Loved Tom Gordon" protagonist: Trisha McFarland, turns to, Tom Gordon for comfort and guidance)
        Triple: ("The Girl Who Loved Tom Gordon", explores themes of, "isolation, fear, and the power of imagination")
        Triple: ("The Girl Who Loved Tom Gordon", was, a critical and commercial success)
        Triple: ("The Girl Who Loved Tom Gordon", has been adapted into, a comic book)
        Triple: ("The Girl Who Loved Tom Gordon", has been adapted into, a stage play)
        \texttt{<Response>\{verified\_response\}}
        \\ \hline
        \textbf{\centering Verification} & Below are the knowledge Triples you extracted based on the text, but there are still some errors in it. For example,  the semantics of the triple are different from the semantics of the corresponding part in the original text or there is a pronoun in the triple. Please check and correct.\newline
        \texttt{<Response>\{verified\_response\}} \newline
        \texttt{<Triples>\{triples\}}. \newline
        Please output all corrected triples directly, including changed and unmodified ones. Don't output any other words.
        \\ \hline
    \end{tabularx}
    }
    \caption{Details of prompt templates in triple extraction and verification.}
    \label{tb:1}
\end{table*}

\section{B Details of Reverse Verification via Triple Reconstruction Prompt Templates}
Table \ref{tb:2}, Table \ref{tb:3}, and Table \ref{tb:4} show the prompts used in the reverse verification strategy mentioned in Sec. 3.4. Table \ref{tb:2} specifically shows the prompts used in the head entity with question answering task, divided into prompts for generating reconstructed questions and prompts for answering those reconstructed questions. Table \ref{tb:3} shows the prompts used in the relation regeneration task. It includes prompts for the model to predict the masked relation in a triple (It uses a one-shot setup.) and prompts to compare if the model's new prediction matches the masked relation. Table \ref{tb:4} shows the prompts used in the fact triple selection based on tail entity task. It includes prompts for the model to replace the tail entity in a triple with another entity from the same category, and prompts for the model to select the factually correct triple from multiple alternatives. Both prompts use a one-shot setup.
\begin{table*}[ht]
    \centering
    \scalebox{0.9}{
    \begin{tabularx}{\textwidth}{>{\centering\arraybackslash}c | X}
        \hline
        \textbf{\centering Prompt Type} & \textbf{\centering Prompt Templates} \\ \hline
        \textbf{\centering Question Generation} & I will give you a set of "Fact triples" and a "Verification triple". You should use all these "Fact triples" to generate a question about "Verification triple", and the answer to the generated question is the head entity of the "Verification triple". Do not include the head entity in your question.
Please pay attention to the output format. Just output the generated questions directly. Do not output any additional words or characters.
Here is an examples:\newline
\texttt{<Input>}
Fact triples:(Paris, is, the capital of France)\newline
Triplet: (Paris, possession, Eiffel Tower)
Verification triple:(Paris, possession, long history)
\texttt{<Output>}
Which city has a long history, the Eiffel Tower, and is the capital of France?\newline
\texttt{<Input>}
\texttt{\{fact triples\}}
\texttt{\{verified triple\}}
        \\ \hline
        \textbf{\centering Question Answering} & You should answer the following question as short as possible. \newline
        \texttt{\{Reconstructed Question\}}
        \\ \hline
    \end{tabularx}
    }
    \caption{Details of prompt templates in Head Entity with Question Answering Task.}
    \label{tb:2}
\end{table*}

\begin{table*}[ht]
    \centering
    \scalebox{0.9}{
    \begin{tabularx}{\textwidth}{c | X}
        \hline
        \textbf{Prompt Type} & \textbf{Prompt Description} \\ \hline
        \textbf{Relationship Prediction} & In the knowledge graph, knowledge triples are a basic data structure used to represent and store information. A knowledge triplet usually consists of three parts: head entity, relationship, and tail entity. This structure helps represent the relationships between entities in a structured way.\newline
        Now given a knowledge triplet, in which the relationship part is replaced by “mask”, you need to predict the relation that the "mask" represents based on the provided information and your knowledge. \newline
        \texttt{<Input>}
        \texttt{Provide information:\{fact triples\}} \newline
        \texttt{Masked triple:\{masked triple\}} \newline
        Please output the complete triple without any additional words.
        \\ \hline
        \textbf{Consistency Comparison} & Please judge whether the following two knowledge triples are semantically consistent and describe the same fact. If so, please answer "yes" directly, otherwise answer "no".\newline
        \texttt{<Input>}
        \texttt{triple1};
        \texttt{triple2}
      \\ \hline
    \end{tabularx}
    }
    \caption{Details of prompt templates in Relation Regeneration Task.}
    \label{tb:3}
\end{table*}

\begin{table*}[ht]
    \centering
    \scalebox{0.9}{
    \begin{tabularx}{\textwidth}{c | X}
        \hline
        \textbf{Prompt Type} & \textbf{Prompt Templates} \\ \hline
        \textbf{Tail Entity Replacement} & I will give you a triple, the format of which is: (Head Entity, Relationship, Tail Entity). Please replace the tail entity in this triplet with another entity of the same type, and provide five such modified triples. Ensure that the semantics of these five triples are different from each other after replacing the tail entity. Start your answer with A: Triple and end with E: Triple. Here is an example:
\texttt{<Input>}(Beijing, capital, China)\newline
\texttt{<Output>}
A:(Beijing, capital, America)
B:(Beijing, capital, England)
C:(Beijing, capital, South Korea)
D:(Beijing, capital, Japan)
E:(Beijing, capital, France)\newline
        \texttt{<Input>}
        \texttt{\{initial triple\}}
        \\ \hline
        \textbf{Answer selection} & The following is a multiple-choice question, with options A, B, C, D, E, and F. Each option corresponds to a knowledge triple. Please think carefully and choose all the knowledge triple that is correctly described and factual based on the provided information and your konwledge, and output the corresponding letter in front of the option. The answer can have more than one option. Make sure your output contains only the letters preceding the option and no other words. Here is an example:\newline
        \texttt{<Input>}
        provide information: (Thomas Alva Edison, invented, the light bulb),(Thomas Alva Edison, is, an entrepreneur). Which of the following options is correct?
A: (Thomas Alva Edison, is, singer)
B: (Thomas Alva Edison, is, actor)
C: (Thomas Alva Edison, is, writer)
D: (Thomas Alva Edison, is, chef)
E: (Thomas Alva Edison, is, entrepreneur)
F: (Thomas Alva Edison, is, inventor) \newline
\texttt{<Output>}EF \newline
\texttt{<Input>}
\texttt{Provide information: \{fact triples\}.\newline
Which of the following options is correct?}
\texttt{\{options\}}
      \\ \hline
    \end{tabularx}
    }
    \caption{Details of prompt templates in Fact Triple Selection based on Tail Entity Task.}
    \label{tb:4}
\end{table*}

\section{C Method of Setting Thresholds}
As mentioned in Sec. 4.1.3, SE, SelfCk-BS, and SelfCk-NLI all return likelihood scores indicating the chance that a sample is a hallucination, rather than labels that specify whether it is a fact or a hallucination. To align with our task, we set thresholds for these three methods in a similar way to our approach. The only difference is that our method returns likelihood scores for a sample being a fact. Specifically, we set the threshold by calculating the mean \(\mu\) and variance \( \sigma^2 \) of the hallucination likelihood scores for all samples obtained by the method in the specific data sets, and then selecting the value from \( \mu\) to \( \mu + 3\sigma\) that maximizes the final metrics as the threshold for the method in the specific data sets.
\section{D Similarity of Representations Between Nodes}
Table \ref{tb:5} presents the full version of results showing the similarity in node distribution in graphs created from two datasets, both with and without the use of RGCN. The similarity between the representations of two nodes is significantly higher after processing with RGCN compared to without RGCN. This suggests that our method incorporates contextual information for each node by utilizing RGCN.
\begin{table*}[htb]
\centering
\begin{tabular}{c|ccccc|ccccc}
\hline
\multirow{2}{*}{\textbf{\centering Methods}} & \multicolumn{5}{c|}{\textbf{PHD}}               & \multicolumn{5}{c}{\textbf{WikiBio}}                      \\
                                  & \textbf{Min} & \textbf{Max} & \textbf{Avg} & \textbf{Std} & & \textbf{Min} & \textbf{Max} & \textbf{Avg} & \textbf{Std} & \\ \hline
RGCN                              & 0.277        & 0.679        & 0.486        & 0.005        & & 0.401        & 0.718        & 0.557        & 0.003        & \\
$-$ RGCN                           & 0.189        & 0.500        & 0.305        & 0.003        & & 0.167        & 0.445        & 0.247        & 0.002        & \\ \hline
\end{tabular}

\caption{Similarity of representations between nodes. RGCN means our full model. $-$ RGCN means comparing without RGCN. (The full version of results)}
\label{tb:5}
\end{table*}

\section{E Significance Test}
To assess whether our method significantly improves performance over the baseline on three datasets, we compare our results to the best-performing baseline, RVQG, using a paired t-test. We find that the difference in F1 scores between our method and RVQG is significant (p-value = 0.049), indicating a notable improvement in F1 score with our method. Additionally, we observe a significant difference in accuracy (p-value = 0.0073), further confirming that our method performs better than RVQG in terms of accuracy. These results demonstrate the effectiveness of our method for detecting hallucinations in long text responses generated by black-box models in zero-resource scenarios.

\section{F Case Study}
In Table \ref{tb:10}, we compare the results of detecting hallucinations in long text responses with other methods. For this response, RVQG and SelfCK-NLI do not identify any factual inaccuracies. SE and SelfCK-BS could detect some hallucinations but missed some of the errors. For example, they overlook the hallucination that needs to consider the dependencies between multiple knowledge facts: \textit{“He was also the first to be born in the French language, as his father had adopted the language and culture of his adopted country”} (the fact is that Honoré IV's native language was French because he was born in Paris, not because his father adopted the language and culture of the country). Conversely, GCA provides a comprehensive and accurate detection result, identifying all factual inaccuracies, since it considers the dependencies between each triple in the response and thoroughly checks each one.
\begin{table*}[htbp]
\centering
\scalebox{0.9}{
\begin{tabularx}{\textwidth}{>{\bfseries}l | X}
\toprule
Methods & \textbf{Results}\\
\midrule
Response & Honoré IV (born Honoré Grimaldi; 14 November 1758 – 3 October 1819) was Prince of Monaco and Duke of Valentinois from 1795 to 1819. He was the son of Prince Honoré III and his wife, Maria Caterina Brignole. He was born in Paris, \textcolor{red}{where his father was living in exile, and was baptized at the church of Saint-Sulpice.} \\
& Honoré IV was a member of the House of Grimaldi, a Genoese family that had ruled Monaco since 1297. He was the first of the Grimaldi to be born in France. \textcolor{red}{He was also the first to be born in the French language, as his father had adopted the language and culture of his adopted country.} \\
& Honoré IV was a strong ruler who sought to modernize Monaco and expand its influence. He was a patron of the arts and sciences, and founded the Monaco Observatory and the Oceanographic Museum of Monaco. He also established the first public library in Monaco. \textcolor{red}{He was a strong supporter of the French Revolution and the Napoleonic Wars, and was a close friend.} \\
\hline
GCA & Honoré IV (born Honoré Grimaldi; 14 November 1758 – 3 October 1819) was Prince of Monaco and Duke of Valentinois from 1795 to 1819. He was the son of Prince Honoré III and his wife, Maria Caterina Brignole. He was born in Paris, \textcolor{blue}{where his father was living in exile, and was baptized at the church of Saint-Sulpice.} \\
& Honoré IV was a member of the House of Grimaldi, a Genoese family that had ruled Monaco since 1297. He was the first of the Grimaldi to be born in France. \textcolor{blue}{He was also the first to be born in the French language, as his father had adopted the language and culture of his adopted country.} \\
& Honoré IV was a strong ruler who sought to modernize Monaco and expand its influence. He was a patron of the arts and sciences, and founded the Monaco Observatory and the Oceanographic Museum of Monaco. He also established the first public library in Monaco. \textcolor{blue}{He was a strong supporter of the French Revolution and the Napoleonic Wars, and was a close friend.}\\
\hline
SE & Honoré IV (born Honoré Grimaldi; 14 November 1758 – 3 October 1819) was Prince of Monaco and Duke of Valentinois from 1795 to 1819. He was the son of Prince Honoré III and his wife, Maria Caterina Brignole. He was born in Paris, \textcolor{blue}{where his father was living in exile, and was baptized at the church of Saint-Sulpice.} \\
& Honoré IV was a member of the House of Grimaldi, a Genoese family that had ruled Monaco since 1297. He was the first of the Grimaldi to be born in France. He was also the first to be born in the French language, as his father had adopted the language and culture of his adopted country. \\
& Honoré IV was a strong ruler who sought to modernize Monaco and expand its influence. He was a patron of the arts and sciences, and founded the Monaco Observatory and the Oceanographic Museum of Monaco. He also established the first public library in Monaco. \textcolor{blue}{He was a strong supporter of the French Revolution and the Napoleonic Wars, and was a close friend.} \\
\hline
SelfCk-BS & Honoré IV (\textcolor{blue}{born Honoré Grimaldi}; 14 November 1758 – 3 October 1819) was Prince of Monaco and Duke of Valentinois from 1795 to 1819. He was the son of Prince Honoré III and his wife, Maria Caterina Brignole. He was born in Paris, \textcolor{blue}{where his father was living in exile, and was baptized at the church of Saint-Sulpice.} \\
& Honoré IV was a member of the House of Grimaldi, a Genoese family that had ruled Monaco since 1297. He was the first of the Grimaldi to be born in France. He was also the first to be born in the French language, as his father had adopted the language and culture of his adopted country. \\
& Honoré IV was a strong ruler who sought to modernize Monaco and expand its influence. He was a patron of the arts and sciences, and founded the Monaco Observatory and the Oceanographic Museum of Monaco. He also established the first public library in Monaco. \textcolor{blue}{He was a strong supporter of the French Revolution and the Napoleonic Wars, and was a close friend.} \\
\hline
SelfCk-NLI &  \textcolor{green}{fact}\\
\hline
RVQG & \textcolor{green}{fact}\\
\bottomrule
\end{tabularx}
}
\caption{Detection results for hallucinations in long text responses using different methods.  \textcolor{red}{red} parts in the Response are hallucinations. \textcolor{blue}{blue} parts in GCA, SE, and SelfCk-BS show the hallucinations detected by these methods. \textcolor{green}{fact} indicates that the method dose not detect the hallucinations and mistakenly judged the response as accurate.}
\label{tb:10}
\end{table*}
\end{document}